\documentclass[letterpaper, 10 pt, conference]{ieeeconf}  

\IEEEoverridecommandlockouts
\usepackage{cite}
\usepackage{amsmath,amssymb,amsfonts}
\usepackage{algorithmic}
\usepackage{graphicx}
\usepackage{textcomp}
\usepackage{xcolor}
\usepackage{booktabs}

\newcommand{\mojtaba}[1]{\textcolor{blue}{#1}}  

\renewcommand{\mojtaba}[1]{#1}

\title{\LARGE \bf
Deep Learning-Enhanced Robotic Subretinal Injection with Real-Time Retinal Motion Compensation
}

\author{Tianle Wu$^{1}$, Mojtaba Esfandiari$^{1}$, Peiyao Zhang$^{1}$, Russell H. Taylor$^{1}$, Peter Gehlbach$^{2}$ and Iulian Iordachita$^{1}$
\thanks{*This work is supported by the U.S. National Institutes of Health under the grants number R01EB023943, R01EB025883, R01EB34397, and
partially by JHU internal funds.}
\thanks{$^{1}$T. Wu, M. Esfandiari, P. Zhang, R. H. Taylor, and I. Iordachita are with the Laboratory for Computational Sensing and Robotics, Johns Hopkins University, Baltimore, MD, USA.}%
\thanks{$^{2}$P. Gehlbach is with the Wilmer Eye Institute, Johns Hopkins Hospital, Baltimore, MD, USA.}%
}

\begin{document}

\maketitle

\begin{abstract}
Subretinal injection is a critical procedure for delivering therapeutic agents to treat retinal diseases such as age-related macular degeneration (AMD). However, retinal motion caused by physiological factors such as respiration and heartbeat significantly impacts precise needle positioning, increasing the risk of retinal pigment epithelium (RPE) damage. This paper presents a fully autonomous robotic subretinal injection system that integrates intraoperative optical coherence tomography (iOCT) imaging and deep learning-based motion prediction to synchronize needle motion with retinal displacement. A Long Short-Term Memory (LSTM) neural network is used to predict internal limiting membrane (ILM) motion, outperforming a Fast Fourier Transform (FFT)-based baseline model. Additionally, a real-time registration framework aligns the needle tip position with the robot’s coordinate frame. Then a dynamic proportional speed control strategy ensures smooth and adaptive needle insertion. Experimental validation in both simulation and \textit{ex vivo} open-sky porcine eyes demonstrates precise motion synchronization and successful subretinal injections. The experiment achieves a mean tracking error below 16.4$\,\mu$m in pre-insertion phases. These results show the potential of AI-driven robotic assistance to improve the safety and accuracy of retinal microsurgery.
\end{abstract}

\section{Introduction}

Age-related macular degeneration (AMD) is the third main cause of visual impairment and blindness worldwide \cite{wong2008natural}. According to a meta-analysis study, the projected number of people with AMD would increase to 288 million in 2040 \cite{wong2014global}. The main current standard for treating AMD is based on symptomatic mitigation by intravitreal injections of anti-vascular endothelial growth factor (VEGF) antibodies, but it has issues such as "resistance to anti-angiogenic therapy" \cite{hubschman2009age} and systemic toxicity \cite{phadke2021review}. Recent advances, such as stem cell transplantation \cite{blenkinsop2012ophthalmologic} and gene therapy \cite{rakoczy2017gene}, offer more effective long-term solutions to treat the cause of AMD, through direct delivery of therapeutic agents into the subretinal space.

Subretinal injection requires cannulation of a microsurgical needle into the retina’s internal limiting membrane (ILM), positioning the needle tip precisely above the retinal pigment epithelial (RPE) cells without damaging the non-regenerative RPE, as it may cause irreversible damage to the eye vision.  
Performing a free-hand subretinal injection is challenging due to the delicacy of RPE, physiological hand tremor (about 182$\,\mu$m \cite{riviere2000study}), small thickness of retina (within the range of about 100 to 570$\, \mu$m \cite{landau1997quantitative}), improper visualization of relative depth of the needle to the target site, to name a few.

Recent advances in robotic technology enable removing hand tremor, providing required positioning accuracy for retinal microsurgeries. Some examples include the Preceyes Surgical
System \cite{van2009design}, the Intraocular Robotic Interventional and Surgical System (IRISS) \cite{rahimy2013robot}, the Steady-Hand Eye Robot (SHER) \cite{uneri2010new}, and others \cite{ida2012microsurgical, nambi2016compact, nasseri2013kinematics}. Two versions of this robot, namely SHER 2.0 and SHER 2.1, are 5 Degrees-Of-Freedom (DOF) serial manipulators that can be used in either cooperative or teleoperation modes, both unimanual \cite{esfandiari2024cooperative} and bimanual \cite{esfandiari2024bimanual} frameworks. 

Human-in-the-loop control modes are limited due to constrained visualization, making accurate positioning of the needle tip inside subretinal space extremely challenging. Several image-guided robot navigation and control strategies are proposed to autonomously guide the robot toward the target area in \textit{ex vivo} porcine eyes using intraoperative optical coherence tomography (iOCT) imaging and deep learning methods \cite{mach2022oct, dehghani2023robotic, zhang2024autonomous, arikan2024real}. Despite promising outcomes in autonomous subretinal injection research, these methods, however, are based on the simplistic assumption that there was no movement in the eye. Neglecting the reciprocating movement of the retina, in actual human eye, caused by respiration, heartbeat, snoring, etc., can deteriorate the needle positioning control accuracy, damaging delicate RPE layer. In a study, the axial movement of retina (along the perpendicular axis to the retina tissue) resulted by heartbeat and respiration is measured as 81$\, \pm \,$3.5$\, \mu$m in amplitude and about 1 Hz in frequency \cite{de2011heartbeat}. Therefore, it is necessary to consider such a reciprocating retina movement in the robot control algorithm to enhance the robotic system's adaptability with real situations and improve patient safety during autonomous robotic subretinal injection procedures.


This paper presents a fully automatic robotic system for subretinal injection, integrating a motion-adaptive control framework with deep learning-based motion prediction. Our approach consists of five structured phases to ensure precise needle positioning despite retinal fluctuations. A U-Net segmentation model extracts the ILM position from iOCT scans, while a Long Short-Term Memory (LSTM) motion prediction module forecasts ILM displacement, outperforming a baseline Fast Fourier Transform (FFT) sine wave model in capturing non-periodic motion patterns. To enhance needle positioning accuracy, we employ a registration step that maps the estimated needle tip position from the image frame to the robot’s local frame, applying noise reduction and a 1D registration transformation for depth alignment. Finally, a proportional speed control algorithm dynamically adjusts needle velocity to synchronize with ILM motion, ensuring smooth and precise insertion. By integrating predictive modeling, adaptive registration, and synchronized control, our method enhances the safety and accuracy of autonomous robotic subretinal injections in dynamic retinal environments. The key contributions are:
\begin{itemize}
    \item Fully automated robotic subretinal injection with real-time motion compensation, adapting to varying retinal motion amplitudes and frequencies in a setting close to clinical conditions. 
    \item A dynamic proportional control strategy for motion synchronization, replacing constant velocity insertion with an adaptive approach that enhances precision.
    \item A robust and efficient LSTM-based motion prediction model, achieving both high accuracy and fast inference with minimal input data. The model significantly outperforms the baseline FFT sinewave approach, enabling real-time prediction within 0.25$\,$s.
    \item Comprehensive validation in both simulation and \textit{ex vivo} porcine eye experiments, demonstrating the system’s effectiveness under ideal conditions and real biological environments. 
\end{itemize}

\section{Methods} \label{sec:methods}

\begin{figure}[t]
    \centering
    \includegraphics[scale=0.17]{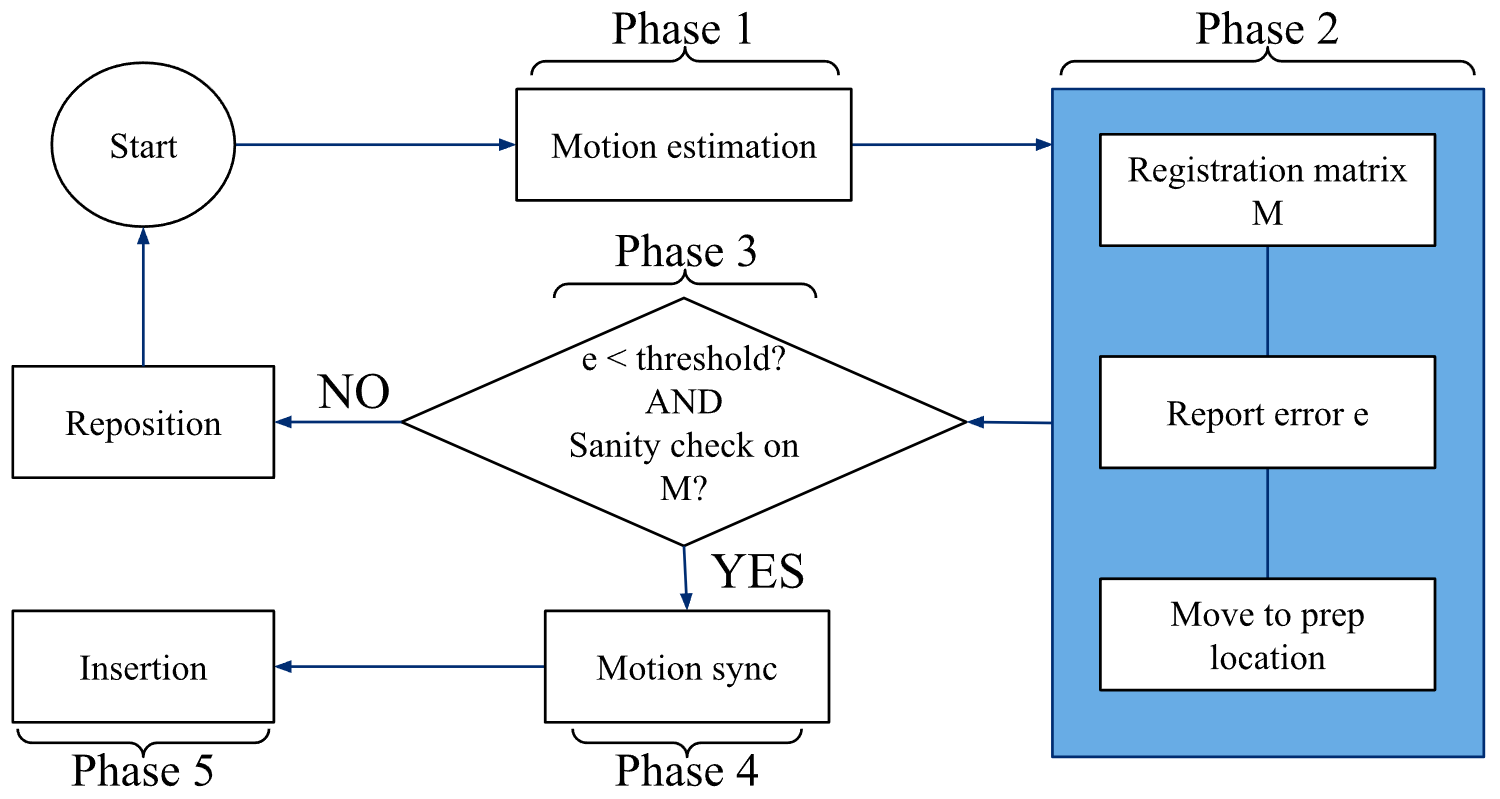}
    \caption{\mojtaba{Flowchart of the proposed deep learning-based autonomous subretinal injection method with retinal motion synchronization and compensation algorithm. }}
    \label{fig:General_Workflow}
\end{figure}

Our control algorithm follows a structured workflow with five distinct phases (Fig.~\ref{fig:General_Workflow}) to ensure precise and synchronized robotic needle insertion. Phase 1: Motion Estimation$-$A segmentation network \cite{arikan2024real} is used to generate the ILM $z$-position. Phase 2: Needle Registration$-$The needle moves to a preparation position, approximately 500$\, \mu$m above the ILM layer. An error report is generated based on the difference between the predicted ILM $z$-position (from Phase 1) and the ground truth ILM $z$-position. The absolute maximum difference is denoted as $e$. A 1D registration process maps the needle tip position from the image frame to the robot’s local frame, generating the registration matrix $M$. Phase 3: Sanity Check$-$The prediction error $e$ is compared against a predefined threshold. Sanity checks on the registration are performed. If it detects the needle in the lower half of the OCT B-scan, which indicates a potential misalignment, the result is a failure in the sanity check. If both checks pass, the process continues; otherwise, manually reposition the needle and restart the process. Phase 4: Motion Synchronization$-$The robot performs synchronized motion while maintaining a fixed distance above the ILM layer. Phase 5: Insertion$-$The robot starts insertion following a predefined control trajectory. Then the fluid is injected to verify successful injection.

The following subsections provide a detailed explanation of the techniques employed in each phase.

\begin{figure}[t]
    \centering
    \includegraphics[scale=0.16]{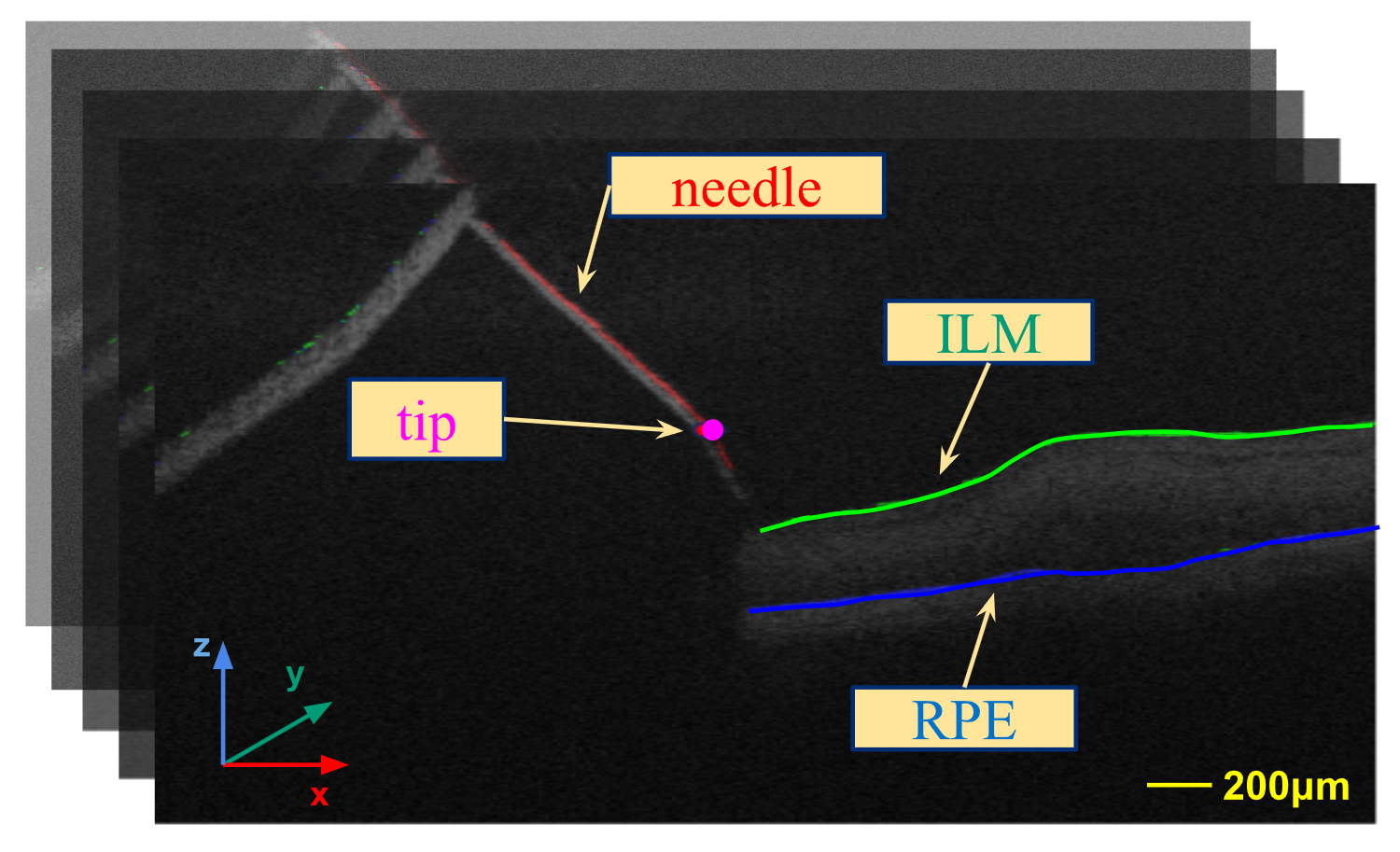}
    \caption{\mojtaba{The surgical needle and the ILM and RPE layers of the retina are segmented using deep learning algorithms and} the B$^5$-scans technique: 5 evenly spaced B-scans \cite{arikan2024real}. \mojtaba{Our proposed method synchronizes the robot velocity with retina's alternating motion along the $z$ axis, providing adaptive synchronized positioning of the needle tip in between ILM and RPE layers without damaging sensitive RPE cells.}  }
    \label{fig:oct_sample}
\end{figure}

\subsection{iOCT Images Processing}
Following the setup in Arikan et al. \cite{arikan2024real}, we capture the retinal depth image using B$^5$-scans (Fig.~\ref{fig:oct_sample}), a compromise between traditional C-scans and single 2D B-scans. It contains five evenly spaced 2D B-scan slices over a 4 × 0.1 mm area. This approach ensures that the needle remains within the imaging field. The images are then passed to a U-Net segmentation model trained on 1707 OCT images collected using the same imaging setup \cite{arikan2024real}. To extract the needle position and retinal layer depth based on the segmentation results, we apply a RANSAC-based line fitting algorithm to remove outlier needle points. The needle tip is identified as the highest point closest to the needle’s center \cite{arikan2024real}.

\subsection{Neural Network-Based ILM Motion Prediction}
To predict the ILM layer's dynamic motion, we implement two models: an LSTM network and a FFT based sine wave predictor as a baseline. The LSTM model is capable of capturing the sequential dependencies in the ILM motion, while the baseline model directly fits a sine wave accordingly.

\subsubsection{LSTM-Based ILM Prediction}
The LSTM model is a recurrent neural network (RNN) designed for time-series forecasting. It consists of a single recurrent layer that captures temporal dependencies in ILM motion, followed by a fully connected (FC) layer that maps the final hidden state to the predicted ILM $z$-position. The architecture follows the model proposed by Hochreiter and Schmidhuber\cite{hochreiter1997long}, and the implementation is done using PyTorch~\cite{paszke2019pytorch}.

\begin{figure}[t]
    \centering
    \includegraphics[scale=0.17]{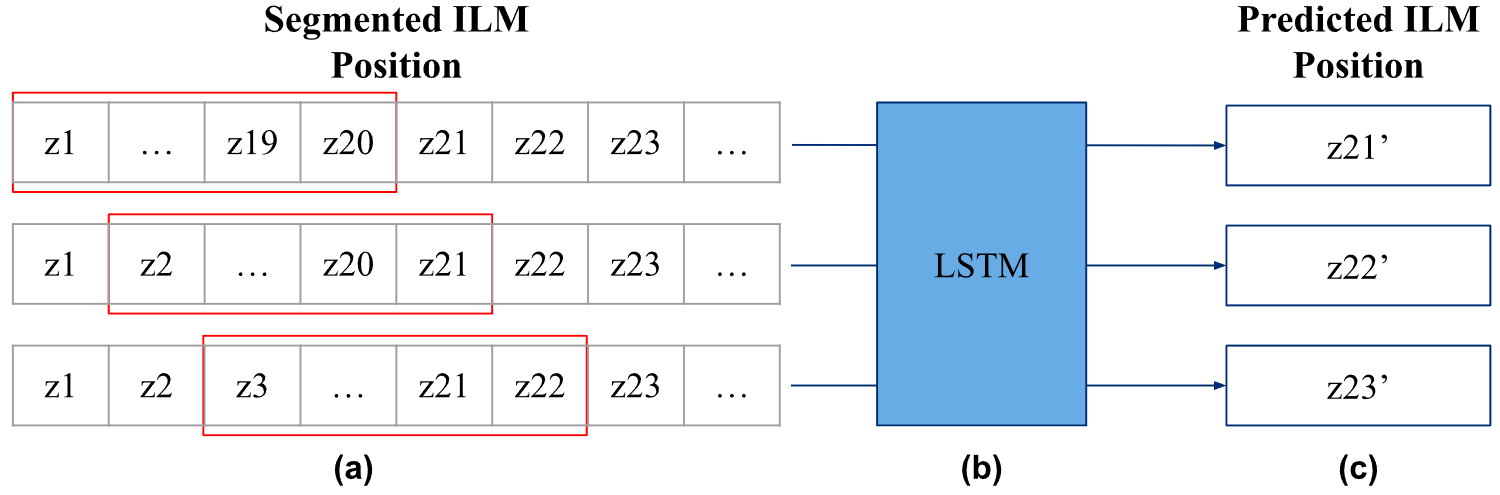}
    \caption{Neural network process to predict ILM position: (a) sequence of segmented ILM $z$-position, given by segmentation results, (b) LSTM network, and (c) one-step-ahead prediction of ILM $z$-position. }
    \label{fig:NNProcess}
\end{figure}

The input to the model is a sequence of 20 consecutive ILM \( z \)-positions (Fig.~\ref{fig:NNProcess}). The model predicts the position at the next time frame. During training, the Mean Squared Error (MSE) Loss function is minimized using the Adam optimizer. The network is trained for 300 epochs with a learning rate of 0.005.

\subsubsection{Baseline Model}
To provide a simple reference for comparison, we implemented an FFT-based sine wave model. This model assumes that the ILM motion follows a periodic pattern. To determine the frequency component of the ILM motion, we perform a Fast Fourier Transform on the input sequence. Then we identify the dominant frequency. The estimated parameters are used to fit a sine wave in the first 20 data points. Future values are predicted based on the fitted function.

\subsubsection{Comparison of Methods}
Both models are evaluated on the same test datasets. The LSTM model is expected to perform more accurately than the baseline because it can adapt to various trends. In contrast, the FFT-based sine wave model is more effective when ILM motion is purely periodic. We use Root Mean Squared Error (RMSE) and Maximum Absolute Error (MaxAE) \mojtaba{to assess performance quantitatively.}

\subsection{Needle Registration}
Accurate needle registration is critical for our system, as the needle position can be noisy while estimated from OCT image segmentation. SHER 2.0 provides relatively precise control, with a translational positioning resolution of 1$\,\mu$m and a bidirectional repeatability of 3$\,\mu$m \cite{uneri2010new}. These features suggest that the needle’s true position in the robot's frame is more reliable than its corresponding location in the OCT image. To improve the control accuracy, we developed a needle registration step to ensure alignment between the image-based tracking and the robot’s positional reference.

\subsubsection{Noise Reduction via Temporal Filtering}
During Phase 1, we keep the needle still and collect 15 consecutive needle position data points obtained from the OCT segmentation results. We apply a two-step filtering approach:
\begin{itemize}
    \item Outlier removal using the Interquartile Range (IQR) method to eliminate extreme values.
    \item Median calculation on the cleaned dataset to obtain a robust estimate of the needle position.
\end{itemize}
The final filtered position in the image frame is then used for registration.

\subsubsection{1D Registration Model}
Since this registration concerns only vertical ($z$ axis) alignment, we applied a 1D transformation model that maps the needle’s detected position from the image frame (pixel coordinates) to the robot frame (millimeter coordinates).

We define the transformation as:
\begin{equation}
    \begin{bmatrix} Z \\ 1 \end{bmatrix} =
    \begin{bmatrix} -b & Z_{\text{init}} + b p_{\text{init}} \\ 0 & 1 \end{bmatrix}
    \begin{bmatrix} p \\ 1 \end{bmatrix}
\end{equation}
where $Z$ is the robot frame coordinate (mm), and $p$ is the needle position in pixel coordinates. $Z_{\text{init}}$ is the robot coordinate of the reference pixel, while $p_{\text{init}}$ is the image coordinate of the reference pixel. The scale factor $b$ (mm per pixel) is derived from the known transformation ratio ($\frac{3.379}{1024}$ mm/px).

\subsection{Needle Control Algorithm}
To synchronize the motion of the needle tip with the ILM layer, we implement a polynomial speed control algorithm in the axial direction:
\begin{equation}
    v = \min(k_v |d_{\text{target}} - d_{\text{current}}|, v_{\max})
\end{equation}
where $d_{\text{current}}$ is the current needle depth, $d_{\text{target}}$ is the desired ILM depth, $k_v$ is the velocity gain factor, $v_{\max}$ is the maximum velocity limit, and $v$ is the computed needle velocity. All parameters are represented in the robot frame (see Fig.~\ref{fig:exp-setup}, left). 

The needle slows down as it nears the ILM layer. The maximum velocity threshold prevents the unsafe speeds. By continuously adjusting the velocity, the system achieves smooth depth synchronization between the needle tip and the ILM layer.

\section{Experiments} \label{sec:experiments}

\subsection{Experimental Setup}

\begin{figure*}[t]
    \centering
    \includegraphics[scale=0.33]{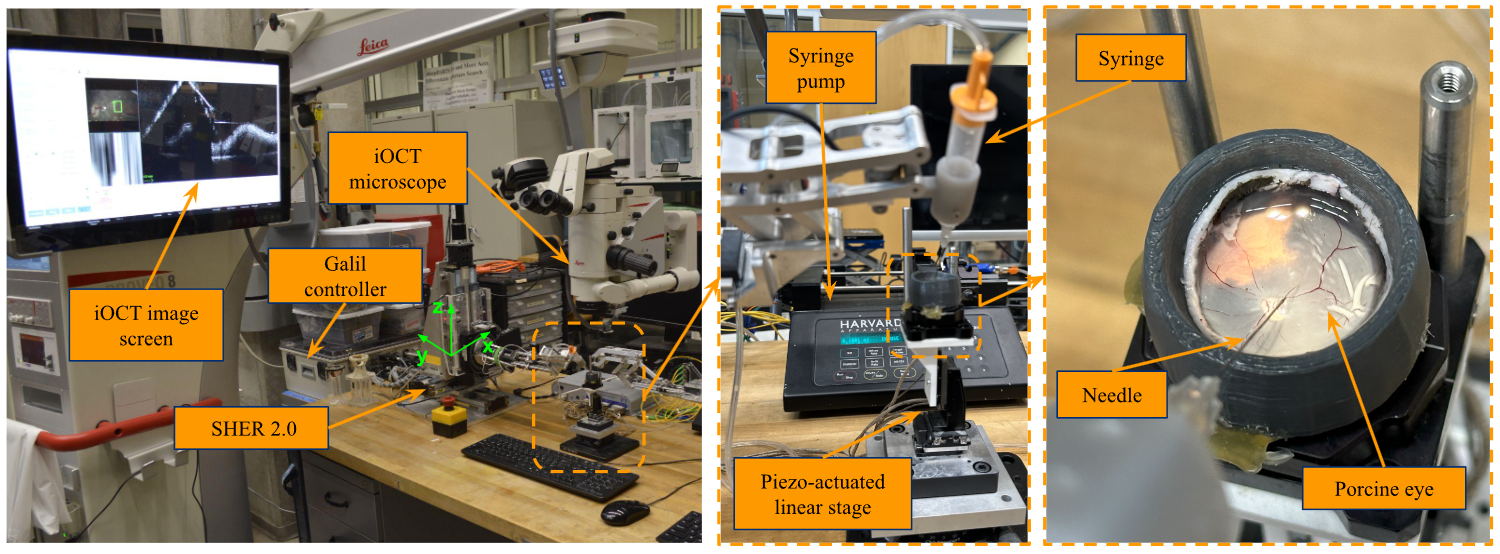}
    \caption{The experimental setup \mojtaba{includes the SHER 2.0, the Galil controller, the Leica iOCT microscope (left), the piezo-actuated linear stage, the syringe pump (middle), a 42-gauge needle, and an open-sky porcine eye (right).} }
    \label{fig:exp-setup}
\end{figure*}

Fig. \ref{fig:exp-setup} shows the experimental setup, comprising the SHER 2.0 and a motion controller (Galil 4088, Galil, Rocklin, CA, USA) for the robot's joint-level velocity control. The setup also includes an iOCT surgical microscope (Proveo 8, Leica Microsystems, Germany), a pump (PHD2000, Harvard Apparatus, USA), a syringe with a 42-gauge needle (INCYTO Co., South Korea) and a piezo-actuated linear stage (Q-Motion Stages, PI Motion and Positioning, MA, USA) to generate the necessary reciprocating motion with desired amplitude and frequencies. The prepared porcine eyes \cite{Pannek2024prepare} are located in a 3D-printed eye holder, which is fixed to the linear stage and moves with it.
The robot's low-level joint velocity controller is developed in C++ utilizing the CISST-SAW libraries \cite{deguet2008cisst}, while the high-level deep learning-based segmentation, prediction, and control algorithms are implemented in Python using OpenCV and PyTorch. All components communicate through the Robot Operating System (ROS) via a TCP-IP connection.

\subsection{Experimental Procedure} \label{sec:experimental_procedure}

\begin{figure}[t]
    \centering
    \includegraphics[scale=0.1]{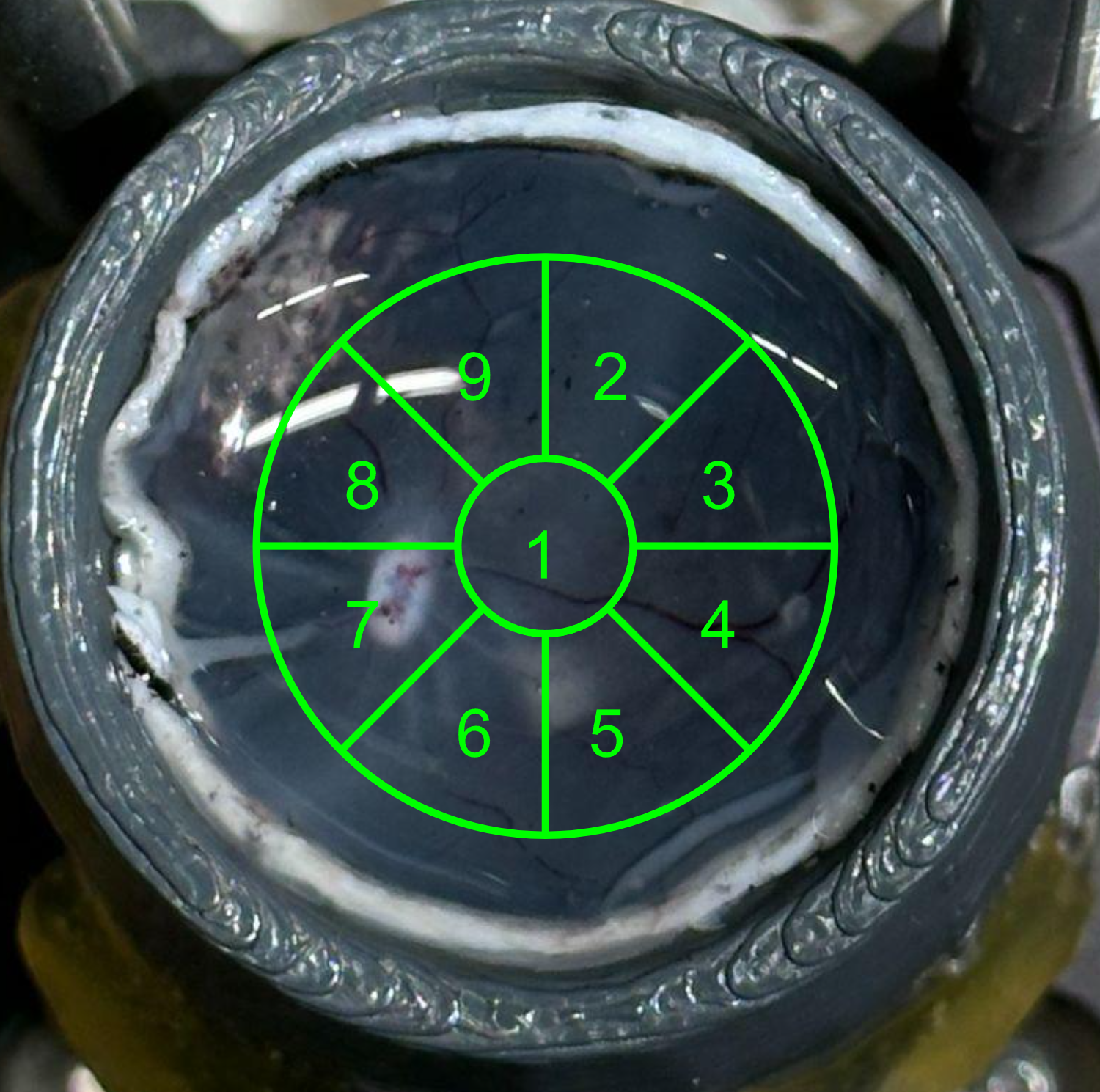}
    \caption{Regions of retina motion collection.}
    \label{fig:retinal_region}
\end{figure}

\begin{figure}[t]
    \centering
    \includegraphics[scale=0.16]{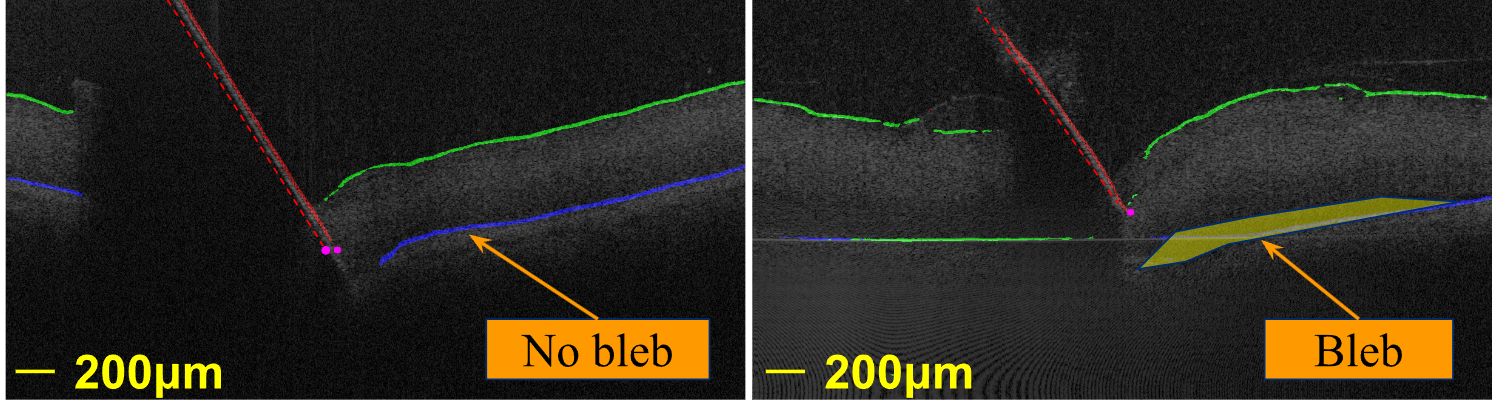}
    \caption{Examples of failed (left) and successful (right) injections in the subretinal area. The yellow area indicates bleb formation. The needle (red) is inserted under the ILM (green) and above the RPE (blue) layer.}
    \label{fig:success_injection}
\end{figure}

\mojtaba{To analyze the performance of our proposed method for autonomous robotic subretinal injection with retinal motion compensation capability}, we \mojtaba{conducted} experiments using \textit{ex vivo} open-sky porcine eyes. The linear stage is programmed to perform sinusoidal motions along the \mojtaba{$z$ axis (see Fig. \ref{fig:oct_sample})} to \mojtaba{simulate alternating retinal motions caused by respiration and heartbeat \cite{de2011heartbeat}.}

We set the sinusoidal amplitude to \( 0.05 \, \)mm, \( 0.1 \, \)mm, and \( 0.15 \, \)mm based on prior research \cite{ourak2019oct}, which indicates that breathing and heartbeat induce periodic retinal motion. According to \cite{ourak2019oct}, breathing motion follows an approximately sinusoidal trajectory, with peak-to-peak amplitudes of $0.1–0.3 \,$mm. Additionally, we set the motion frequency based on clinical conditions. Patients during retinal surgery are typically under anesthesia, where breathing rates range from 6.4 to 19.5 breaths per minute (bpm) \cite{drummond2013breathing}. To evaluate our system’s robustness across different breathing rates, we \mojtaba{tested on different} frequencies of 8, 9, and 10$\,$bpm. 

A higher frequency increases the difficulty of motion tracking. Successful performance at higher frequencies typically implies robustness at lower ones. The breathing rate in bpm is converted to frequency in Hertz (Hz) using:
\begin{equation}
    f_{\text{Hz}} = \frac{\text{bpm}}{60}
\end{equation}
and is used as the input command to the \mojtaba{piezo-actuated} linear stage.

\subsubsection{Retinal Layer Motion Data Collection}
To collect retinal depth data, we \mojtaba{used} 12 porcine eyes, each divided into 9 distinct regions. A random region was selected for tracking, as shown in Fig.~\ref{fig:retinal_region}. However, some eyes \mojtaba{had} corrupted or low-quality OCT images, making them unsuitable for analysis. Ultimately, we \mojtaba{gathered} 18 valid datasets from 8 eyes, totaling 7,200 position data points of both ILM and RPE layer.

The data sampling rate is 4$\,$Hz (i.e., four segmentation results per second). The depth measurements range from 1.68$\,$mm to 2.7$\,$mm, relative to the OCT imaging frame.

\subsubsection{Needle Insertion Experiments}
We \mojtaba{conducted} insertion experiments in both a simulation environment and \textit{ex vivo} porcine eyes to assess the accuracy of our control algorithm.

The simulated motion trajectory closely \mojtaba{followed} a pure sinusoidal path, with 0.1$\,$mm amplitude and 8 bpm frequency. It provides idealized conditions to evaluate the precision of the control algorithm without external biological disturbances.

For \textit{ex vivo} experiments, we \mojtaba{performed} insertions with 0.05$\,$mm amplitude and 8$\,$bpm frequency. Unlike simulation, \textit{ex vivo} testing \mojtaba{introduced} biological uncertainties that impact motion tracking, including:
\begin{itemize}
    \item \textbf{Tissue \mojtaba{deformation after insertion}:} Retinal layers show slight shifts due to the needle insertion. The relative positions of the ILM and RPE layers change.
    \item \textbf{Segmentation \mojtaba{noise in image-based tracking}:} During insertion, the needle partially blocks the OCT image, which lead to reduced segmentation reliability.
\end{itemize}

\subsubsection{Success Criteria for Subretinal Injection}
To evaluate the success of subretinal injections, we visually \mojtaba{inspected} post-injection retinal \mojtaba{reactions}. Fig.~\ref{fig:success_injection} presents examples of both successful and failed injections. A successful injection \mojtaba{was} identified by the formation of a stable bleb under the retina.

\section{Results and Discussion} \label{sec:results}

This section discusses our proposed method for autonomous robotic subretinal injection with a retinal motion compensation algorithm based on an LSTM network. The algorithm is tested in both simulation environment and \textit{ex vivo} open-sky porcine eyes. The performance of the LSTM network in predicting the retina's ILM $z$-position is compared with a baseline FFT model.

\subsection{Retina Motion Prediction Network}
\begin{figure}[t]
    \centering
    \includegraphics[scale=0.17]{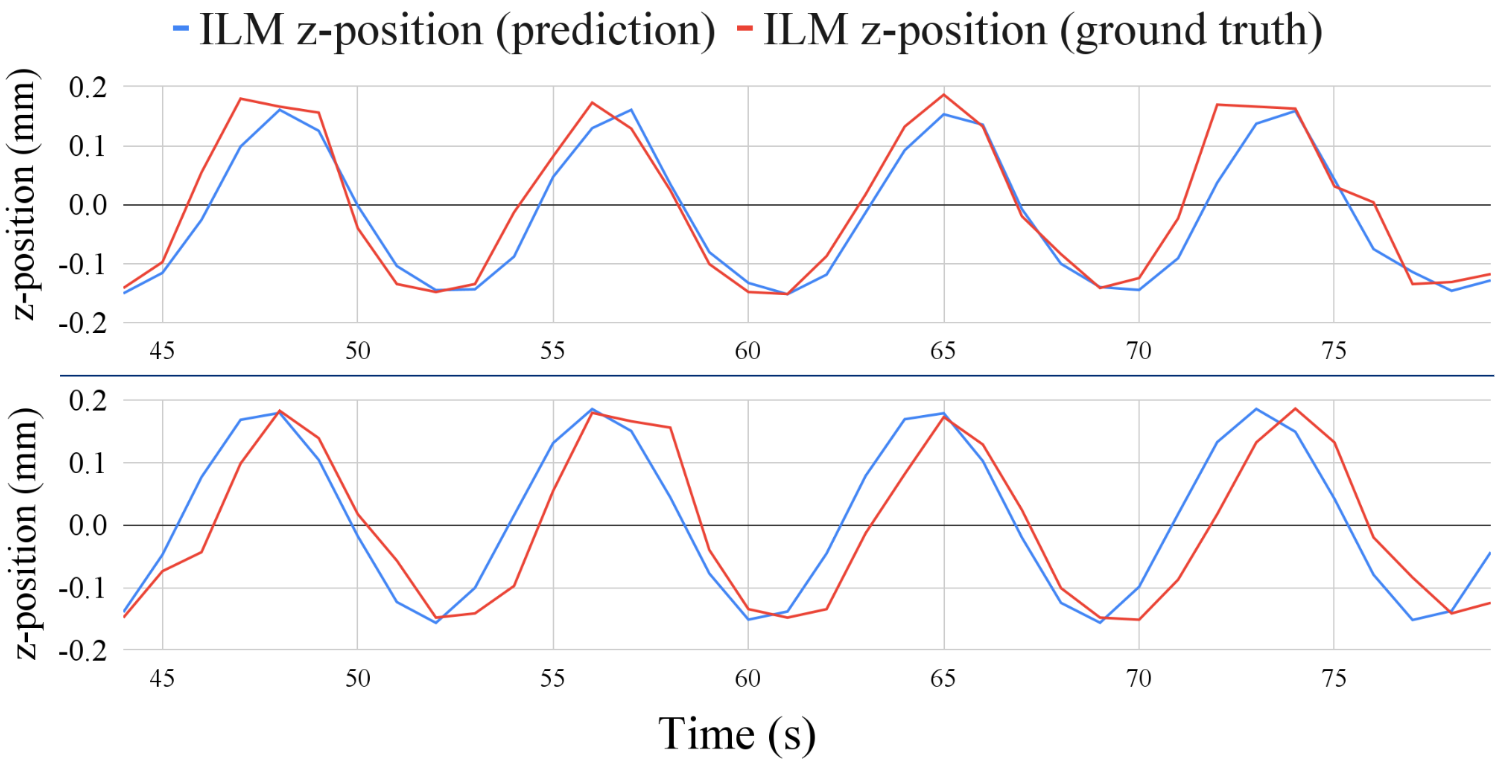}
    \caption{An example of prediction of ILM $z$-position  vs. ground truth using LSTM (top) and FFT (bottom) methods for about four reciprocating periods.  }
    \label{fig:lstm_fft_visual}
\end{figure}

\begin{table}[t]
    \centering
    \caption{LSTM Performance Comparison}
    \begin{tabular}{lccc}
        \toprule
        Amplitude (mm) & Frequency (bpm) & RMSE ($\mu$m) & MaxAE ($\mu$m) \\
        \midrule
        0.05 & 8 & 6.69 & 18.21 \\
        0.05 & 9 & 7.70 & 25.55 \\
        0.05 & 10 & 10.20 & 37.03 \\
        \midrule
        0.1 & 8 & 15.92 & 53.86 \\
        0.1 & 9 & 17.33 & 63.05 \\
        0.1 & 10 & 25.03 & 90.22 \\
        \midrule
        0.15 & 8 & 24.46 & 75.78 \\
        0.15 & 9 & 24.91 & 90.67 \\
        0.15 & 10 & 33.73 & 137.18 \\
        \bottomrule
    \end{tabular}
    \vspace{0.1cm}
    \label{tab:lstm_rmse_me}
\end{table}

\begin{table}[t]
    \centering
    \caption{FFT vs. LSTM Performance Comparison}
    \label{tab:fftvslstm_rmse_me}
    \begin{tabular}{llcccc}
        \toprule
        Amp.(mm) & Freq.(bpm) & \multicolumn{2}{c}{RMSE ($\mu$m)} & \multicolumn{2}{c}{MaxAE ($\mu$m)} \\
        \cmidrule(lr){3-4} \cmidrule(lr){5-6}
        & & LSTM & FFT & LSTM & FFT \\
        \midrule
        0.05 & 8  & 6.69  & 15.88  & 18.21  & 38.35 \\
        0.15 & 10 & 33.73 & 134.89 & 137.18 & 287.22 \\
        \bottomrule
    \end{tabular}
    \vspace{0.1cm}

\end{table}

We present a detailed performance analysis of the proposed LSTM model, compared to the baseline FFT sine wave model. We select RMSE and MaxAE as the evaluation metrics. Based on the previous 20 time frames of retinal layer movement ($z$-position), the LSTM network predicts the ILM $z$-position at the next time frame (one-step-ahead prediction of approximately 0.25$\,$s).

RMSE measures the average magnitude of prediction errors and penalizes larger errors. MaxAE captures the largest prediction error in the dataset, reflecting the worst-case scenario. This is important to evaluate the reliability of real-time predictions in surgical procedures.

Table~\ref{tab:lstm_rmse_me} summarizes the performance of LSTM predictions across the nine datasets. The LSTM model achieves consistently low errors, with the lowest error observed at Amp=0.05$\,$mm, Freq=8$\,$bpm, where RMSE is 6.69$\,\mu$m and MaxAE is 18.21$\,\mu$m. The highest error occurs at Amp=0.15$\,$mm, Freq=10$\,$bpm, with RMSE of 33.73$\,\mu$m and MaxAE of 137.18$\,\mu$m. We also observe that larger amplitudes and frequencies in our dataset lead to higher errors (both RMSE and MaxAE). This is intuitive, as higher variations in movement introduce more noise and unpredictable factors, making it harder for the model to capture consistent patterns.

To evaluate the effectiveness of our LSTM model, we compare its RMSE and MaxAE against the FFT sine wave model in the best and worst cases. Table~\ref{tab:fftvslstm_rmse_me} presents this comparison. The LSTM model outperforms the FFT model in both cases, achieving lower RMSE and MaxAE in each instance. This result confirms that our LSTM approach is more effective than the FFT sine wave model in general.

To provide further insights, Fig.~\ref{fig:lstm_fft_visual} presents a zoomed-in comparison of LSTM and FFT model predictions against the ground truth. The plot illustrates that the LSTM model closely follows the actual data trend. However, the FFT sine wave model shows latency in its predictions.

\subsection{Insertion Quantitative Performance}
\begin{figure*}[t]
    \centering
    \includegraphics[scale=0.335]{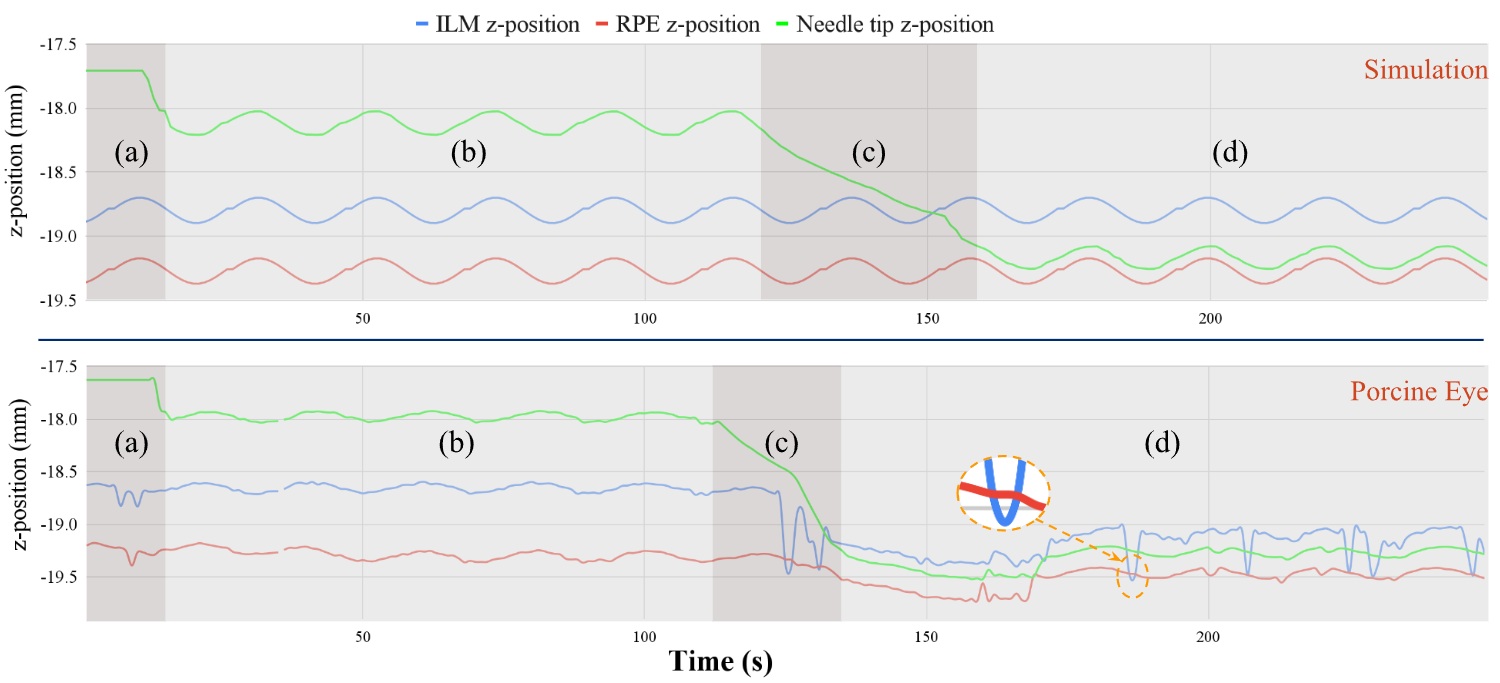}
    \caption{Examples of automatic insertion performance: simulation (top) vs. actual \textit{ex vivo} porcine eye (bottom). (a) Phases 1$-$3: Robot performs motion estimation and moves the needle to the preparation location. (b) Phase 4: Robot begins motion synchronization above the ILM layer at 675$\,\mu$m. (c) Phase 5: Robot initiates needle insertion. (d) Robot executes an additional motion synchronization within the retinal layers. The zoomed area around 180\,s (dashed circle) highlights an example of segmentation noise caused by tissue deformation, as the ILM (blue) appears lower than the RPE (red), which is unlikely in a real scenario.}
    \label{fig:insertion_data_plot}
\end{figure*}

\begin{table}[t]
    \centering
    \caption{Comparison of Control Precision in Simulation and \textit{ex vivo} porcine Eye Experiments. (all units are in $\mu m$)}
    \begin{tabular}{lcccc}
        \toprule
        Condition & Offset & RMSE & MaxAE & Mean Error \\
        \midrule
        \textbf{Simulation} \\ (Above ILM) & 675.8  & 12.61  & 23.0  & 10.8  \\
        \textbf{Simulation} \\ (Inside Retina) & 101.37 & 9.0  & 22.3  & 7.5  \\
        \midrule
        \textbf{\textit{ex vivo} Eye} \\ (Above ILM) & 675.8  & 16.4  & 43.5  & 13.8  \\
        \textbf{\textit{ex vivo} Eye} \\ (Inside Retina) & 202.74 & 36.25  & 200.0  & 22.6  \\
        \bottomrule
    \end{tabular}
    \label{tab:control_comparison}
\end{table}

We compare the control performance in the simulation environment to an \textit{ex vivo} porcine eye experiment. The precision of our robotic motion control in motion synchronization and needle insertion phases is reported. The comparison provides insights into how well the system performs in an ideal environment (simulation) versus a realistic biological setting (\textit{ex vivo} eye).

Fig.~\ref{fig:insertion_data_plot} presents the insertion process in two different environments: simulation (top) and \textit{ex vivo} porcine eye (bottom). The insertion procedure is divided into several phases. To quantify the control precision, we analyze RMSE, MaxAE, and Mean Error values in two critical phases:
\begin{itemize}
    \item Motion synchronization above the ILM layer (relative to ILM).
    \item Needle tracking within the retinal layers (relative to RPE).
\end{itemize}
Table~\ref{tab:control_comparison} summarizes the error metrics for both simulation and \textit{ex vivo} experiments.

The results indicate that RMSE and MaxAE values are lower in both synchronization and insertion phases in the simulation compared to the \textit{ex vivo} experiment. The system follows smooth motion trajectories and maintains high accuracy. However, since the simulation lacks biological complexities such as tissue deformation, these results can only be evaluated as an idealized benchmark.

In contrast, the \textit{ex vivo} porcine eye experiment introduces biological uncertainties. The result shows a noticeable increase in RMSE and MaxAE values. Particularly in Phase 5 (post-insertion), RMSE increases from 9$\,\mu$m (simulation) to 36.25$\,\mu$m, and MaxAE spikes from 22.3$\,\mu$m to 200$\,\mu$m. This decrease in accuracy suggests that the insertion process is influenced by tissue interactions and potential segmentation noise, as previously discussed in Sec.~\ref{sec:experimental_procedure}. We conduct 8 injection trials, with 5 successfully resulting in retinal detachment formation and a stable bleb appearance.

\section{Future Work}
One area for improvement is motion control optimization. The current dynamic proportional control method is sufficient only for small velocity variations. More advanced control strategies, such as adaptive learning-based controllers, could be implemented. These approaches could compensate sudden retinal motion variations and adjust the insertion trajectory in real time.

Another improvement can be applied to image segmentation for real-time retinal layer and needle tracking. The current method lacks accuracy when the needle is inside the retina. In the future, we will fine-tune the model and train with a more comprehensive dataset. 

Finally, further validation is needed in intact porcine eye experiments. Currently our current \textit{ex vivo} studies use an open-sky setup, which does not fully replicate the surgical environment. The intact porcine eye model will provide a more clinically relevant situation for evaluating the system’s performance.

\section{Conclusion} \label{sec:conclusion}
This paper presented a fully autonomous robotic system for subretinal injection, integrating real-time motion compensation using iOCT imaging and deep learning-based motion prediction. \mojtaba{Our proposed method enables an autonomous needle insertion into the subretinal space with synchronized retinal motion compensation.} \mojtaba{This is achieved} by incorporating an LSTM-based ILM motion predictor, a registration framework for accurate needle positioning, and a dynamic proportional speed control algorithm. Experimental validation in both simulation \mojtaba{environment} and \textit{ex vivo} porcine eyes demonstrated the capability of the \mojtaba{proposed method} \mojtaba{in real-time tracking of retinal motions}, \mojtaba{synchronizing the robotic surgical instrument with the retina motion}, \mojtaba{and performing safe} subretinal injections.


\bibliographystyle{IEEEtran}
\bibliography{references}

\end{document}